%% file: colm2026_conference.tex
\documentclass{article} 
\usepackage[final]{colm2026_conference}

\usepackage{microtype}
\usepackage{hyperref}
\usepackage{url}
\usepackage{booktabs}

\usepackage{lineno}

\definecolor{darkblue}{rgb}{0, 0, 0.5}
\hypersetup{colorlinks=true, citecolor=darkblue, linkcolor=darkblue, urlcolor=darkblue}

\usepackage{amsmath}
\usepackage{amssymb}
\usepackage{mathtools}
\usepackage{amsthm}

\usepackage[most]{tcolorbox} 
\usepackage{graphicx}       
\usepackage{caption}        
\usepackage{calc}           

\newlength{\sbsGap}
\newlength{\sbsImgWi}\newlength{\sbsImgHi}
\newlength{\sbsImgWii}\newlength{\sbsImgHii}
\newlength{\sbsTargetH}
\newlength{\sbsWi}\newlength{\sbsWii}
\newlength{\sbsWiiAtHi}\newlength{\sbsTotalAtHi}
\newlength{\sbsAvail}

\title{BoT-GRPO: Efficient Process-Reward RL for Reasoning via Bag-of-Token Aggregation}

\author{%
Yingxiang Yang, Weihang Xiao \& Zhunxuan Wang \\
Amazon AGI \\
\texttt{\{yayingxi,weihanx\}@amazon.com, wzhunxua@amazon.co.uk} \\
\And
Joshua Flashner\thanks{Work done during an internship at Amazon.} \\
Virginia Tech \\
\texttt{jflashner@vt.edu} \\
\AND
Niresh Agarwal \\
Amazon AGI \\
\texttt{nirea@amazon.com}
}

\begin{document}

\ifcolmsubmission
\linenumbers
\fi

\maketitle

\begin{abstract}
Reinforcement learning is now central to eliciting reasoning in large language models, while in the popular algorithm Group Relative Policy Optimization (GRPO) every token in a rollout receives the same advantage. We ask how to make process supervision \emph{efficient}: accelerating convergence and improving final quality without the cost of value networks. We propose Bag-of-Tokens Group Relative Policy Optimization (BoT-GRPO), which extends GRPO to token-level reward models through a length-invariant ``bag of tokens'' aggregation: it collects all token-level rewards across rollouts, weights each by the inverse of its source sequence length, and computes per-token advantages relative to weighted group statistics. BoT-GRPO is critic-free, and is a drop-in replacement wherever GRPO is used when token-level reward is available. On React front-end code generation, BoT-GRPO reaches $80\%$ compile rate up to $1.9\times$ faster than GRPO and converges faster than modern GRPO variants (GSPO, DAPO, PURE) while reaching higher final compile and VLM-judged win rates. On a second task, AIME mathematical reasoning, BoT-GRPO delivers absolute Pass@$k$ gains up to 8.1\% over GRPO in half the steps. For both tasks we compare the algorithm's performance on reasoning vs. non-reasoning base-model families (Qwen2.5-3B, SmolLM3-3B, Phi-4-mini-reasoning). Our experiments also yield a practical recipe for the reward model itself: reward \emph{stability} matters more than richness: clean, bounded, stable fine-grained signals consistently accelerate learning where noisier alternatives stall.
\end{abstract}

\input{intro}

\input{related_work}

\input{method}

\input{reward}

\input{experiments}

\input{conclusion}

\bibliography{references}
\bibliographystyle{colm2026_conference}

\input{appendix}

\end{document}

%% file: intro.tex
\section{Introduction}

Reinforcement learning with verifiable rewards (RLVR) has become the dominant recipe for eliciting reasoning in large language models, from DeepSeek-R1 to open post-training pipelines~\citep{deepseek2025r1, lambert2024rlvr}. At the center of this recipe is Group Relative Policy Optimization (GRPO), which discards PPO's value network and instead normalizes rewards within a group of sampled rollouts, trading a learned critic for cheaper, more stable optimization~\citep{shao2024deepseekmath}. Yet the compute bill for RL post-training remains dominated by the many rollouts needed to see reward signal, and GRPO uses that signal inefficiently: because advantages are computed once per sequence, every token in a rollout is updated with the identical, sequence-level signal regardless of whether it helped or hurt. This coarse-grained credit assignment spends gradient uniformly across long chains of thought, slowing convergence and inflating the number of rollouts, GPU-hours, and wall-clock time required to reach a target quality~\citep{liu2025drgrpo}.

Denser, better-targeted reward is the natural remedy, and process reward models (PRMs) that score intermediate steps demonstrably outperform outcome-only supervision~\citep{lightman2023prm}. But process supervision has historically reintroduced exactly the costs that make RLVR expensive. Monte-Carlo alternatives estimate step values by rolling out many continuations per step~\citep{wang2024mathshepherd}; verifier and reward models require their own training pipelines~\citep{wu2025opv}; TD-learning methods bring back a critic~\citep{xi2025agentprm}; and sequence-similarity approaches depend on curated expert trajectories~\citep{deng2025srl}. Even implicit-PRM methods that obtain token-level rewards more cheaply~\citep{yuan2024implicitprm, cui2025prime} still change the reward source rather than the way GRPO turns per-token rewards into advantages. The question for \emph{efficient} reasoning is thus not only what to reward, but how to fold token-level signals into GRPO so that they \emph{accelerate} convergence without adding a value network, an auxiliary model to train, or per-token cost that grows with the length of reasoning.

We propose \textbf{Bag-of-Tokens Group Relative Policy Optimization (BoT-GRPO)}, a critic-free extension of GRPO that consumes any external token-level reward and assigns per-token advantages through a length-invariant ``bag of tokens'' aggregation. Weighting each token by the inverse of its source-sequence length lets every rollout contribute equally to the group statistics regardless of how long its reasoning ran, correcting the length bias that plagues sequence-level GRPO~\citep{liu2025drgrpo} while adding only bookkeeping overhead. BoT-GRPO makes no assumption about where the token reward comes from: any combination of a global outcome and a bounded local signal works. Fine-grained credit assignment thus becomes a lever on training efficiency rather than a separate training system, keeping the whole procedure within reach of a single commodity GPU at the model scales we study.

We evaluate BoT-GRPO on both front-end code generation and AIME math problems. For front-end code generation, correctness (does it compile?) and quality (does it look good?) are both measured, and we verify that the findings transfer to AIME mathematical reasoning. Against standard GRPO and three modern GRPO variants aimed at stability and credit assignment (GSPO~\citep{zheng2025gspo}, DAPO~\citep{yu2025dapo}, PURE~\citep{cheng2025pure}), BoT-GRPO reaches target quality in markedly fewer steps and attains higher final compile and win rates, with the speedup holding across three base-model families. We see similar faster convergence and higher stable performance for the AIME use case as well.

\looseness=-1
Our contributions are: (1)~\textbf{BoT-GRPO}, a critic-free, drop-in token-level advantage estimator that turns external local rewards into a convergence-speed lever without extra models or unbounded cost; a length-invariant $1/L_k$ weighting additionally corrects the response-length bias of sequence-level GRPO~\citep{liu2025drgrpo}. (2)~Evidence that on front-end code generation BoT-GRPO converges faster and reaches higher quality than GRPO and three modern variants (GSPO, DAPO, PURE) (3)~A practical \textbf{minimal reward recipe} for code generation, identifying which reward signals are necessary versus optional and how to keep them stable. (4)~Supporting AIME results and ablations on reward signal designs. (5) An empirical comparison of BoT-GRPO's behavior across reasoning and non-reasoning base models.

%% file: related_work.tex
\section{Related Work}

\paragraph{Efficient RL for Reasoning.}
RLVR post-training now underpins state-of-the-art reasoning models~\citep{deepseek2025r1, lambert2024rlvr}, but its cost is dominated by the rollouts needed to observe reward, making sample and compute efficiency a first-order concern. GRPO removes PPO's value network by normalizing rewards within a sampled group~\citep{shao2024deepseekmath}, yet analyses of R1-Zero-style training show that its sequence-level objective induces response-length and difficulty biases that waste optimization and inflate reasoning length~\citep{liu2025drgrpo}. A complementary line targets inference-time efficiency, curbing the ``overthinking'' of long chains of thought or learning to control reasoning length directly with RL~\citep{chen2024overthinking, aggarwal2025l1}. BoT-GRPO instead attacks \emph{training-time} efficiency: its length-invariant aggregation both corrects the length bias and densifies credit assignment so that fewer rollouts are needed to reach a target quality.

\paragraph{Process Rewards and Their Cost.}
\looseness=-1
Process reward models that score intermediate steps outperform outcome-only supervision~\citep{lightman2023prm}, but the denser signal has historically been expensive: step labels were human-annotated~\citep{lightman2023prm}, Monte-Carlo estimation rolls out many continuations per step~\citep{wang2024mathshepherd}, and verifier models require separate training pipelines~\citep{wu2025opv}. Recent work lowers this cost by deriving process rewards implicitly from an outcome-trained model, either offline~\citep{yuan2024implicitprm} or online during RL~\citep{cui2025prime}. A parallel thread focuses on the \emph{design} of the step-level signal itself rather than its cost: StepWiser~\citep{xiong2025stepwiser} recasts step evaluation as a generative meta-reasoning task, training an RL judge that reasons about each step and explains its verdict from relative rollout outcomes rather than emitting a bare label, which improves both training-time supervision and inference-time search. All of these methods focus on \emph{producing} a better or cheaper token-level signal $S_t^{(k)}$; BoT-GRPO is orthogonal to that choice, addressing instead how any such reward -- generative-judge, rule-based, or implicit -- is \emph{aggregated} into per-token advantages fairly across variable-length rollouts.

\paragraph{Token-Level Reward Methods.}
Several approaches move credit assignment to token-level granularity. Dense reward redistribution transforms sequence rewards into token-level signals using attention weights~\citep{chan2024denseRewardFree}. Inverse-Q* estimates token-wise rewards with PPO-like updates~\citep{xia2024inverseq}, TPPO trains token-wise reward models~\citep{ouyang-etal-2025-token}, and Q-RM learns token-level Q-functions achieving 12$\times$ faster convergence on GSM8K~\citep{pmlr-v267-chen25ca}. BoT-GRPO shares the token-level goal but keeps GRPO's critic-free simplicity, requiring no additional learned reward or value network.

\paragraph{GRPO Variants.}
A growing family of GRPO extensions addresses credit assignment and stability. Entropy-based methods (GTPO~\citep{zhang2025gtpo}, EAPO~\citep{he2026eapo}) weight tokens by policy entropy, hypothesizing that high-entropy positions correspond to cognitively demanding decisions. Eligibility-trace methods redistribute outcome rewards backward: Sullivan and Koller~\citep{sullivan2025grposecret} propose $\lambda$-GRPO to correct imbalanced process steps, while GRPO-$\lambda$~\citep{parthasarathi2025grpolambda} uses token-level log-probabilities as a critic-free TD approximation, and $\lambda$-GRPO~\citep{liu2025lambdagrpo} enables learnable token preferences. For multi-turn settings, GiGPO~\citep{feng2025gigpo} introduces two-level grouping, and Wei et al.~\citep{wei2025turnlevel} study turn-level reward design for search agents. A parallel line improves the GRPO objective itself: DAPO~\citep{yu2025dapo} and VAPO~\citep{wang2025vapo} use gradient normalization, clip-higher, and dynamic sampling; GSPO~\citep{zheng2025gspo} moves importance ratios and clipping to the \textit{sequence} level for more stable large-model training; and PURE~\citep{cheng2025pure} replaces summation with min-form credit assignment to curb reward hacking under process supervision. We take DAPO, GSPO, and PURE as representative modern GRPO variants and compare against them in \S\ref{sec:react_variants}. BoT-GRPO differs from all these along two axes: it accepts \textit{external} token-level rewards from any source rather than deriving weights from the policy or redistributing outcome rewards, and its length-invariant bag-of-tokens aggregation is \textit{orthogonal}, able to incorporate entropy-weighted, eligibility-trace, sequence-level, or turn-level signals as its reward source.

%% file: method.tex
\section{Method}
\label{sec:method}

We consider autoregressive language models where a policy $\pi_\theta$ generates sequences $y = (y_1, \ldots, y_L)$ conditioned on prompts $x$. For each prompt we sample a \emph{group} of $K$ rollouts $\{y^{(1)}, \ldots, y^{(K)}\}$; we write $L_k$ for the token length of rollout $k$, $y_t^{(k)}$ for its $t$-th token, and $y_{<t}^{(k)}$ for the tokens preceding position $t$. A sequence-level reward is denoted $R^{(k)}\in\mathbb{R}$ and a token-level reward $r_t^{(k)}\in\mathbb{R}$. Advantages are written $A^{(k)}$ (sequence-level) or $A_t^{(k)}$ (token-level), $\mathcal{L}$ denotes the per-rollout loss, and $\epsilon>0$ is a small constant for numerical stability. Standard GRPO~\citep{shao2024deepseekmath} normalizes sequence-level rewards within groups but assigns identical advantage signals to all tokens regardless of their individual contributions.

\paragraph{Standard GRPO.} With group mean $\mu=\tfrac{1}{K}\sum_j R^{(j)}$, omitting KL divergence and clipping:
\begin{equation}
A^{(k)} = \frac{R^{(k)} - \mu}{\sqrt{\tfrac{1}{K}\sum_{j}(R^{(j)} - \mu)^2} + \epsilon}, \qquad
\mathcal{L}_{\text{GRPO}}^{(k)} = -\frac{1}{L_k} \sum_{t=1}^{L_k} A^{(k)} \log \pi_\theta(y_t^{(k)} | x, y_{<t}^{(k)})
\end{equation}
\paragraph{BoT-GRPO.} We extend GRPO to token-level rewards via a ``bag of tokens'' aggregation: (1)~collect all tokens across rollouts into a single pool, (2)~weight each by $w_k = 1/L_k$ (inverse source sequence length), (3)~compute weighted group statistics, (4)~calculate per-token advantages relative to these statistics.

The weighted mean and variance are:
\begin{equation}
    \mu = \frac{1}{K}\sum_{k=1}^K\frac{1}{L_k}\sum_{t=1}^{L_k}r_t^{(k)}, \qquad \sigma^2 = \frac{1}{K}\sum_{k=1}^K\frac{1}{L_k}\sum_{t=1}^{L_k}(r_t^{(k)}-\mu)^2\label{eq:weighted_mean_var}
\end{equation}
The token-level advantage and BoT-GRPO loss are:
\begin{equation}\label{eq:token_advantage}
A_t^{(k)} = \frac{r_t^{(k)} - \mu}{\sigma + \epsilon}, \qquad
\mathcal{L}_{\text{BoT-GRPO}}^{(k)} = -\frac{1}{L_k} \sum_{t=1}^{L_k} A_t^{(k)} \log \pi_\theta(y_t^{(k)} | x, y_{<t}^{(k)})
\end{equation}

\paragraph{Length-Invariant Normalization.} The $1/L_k$ weighting ensures each sequence contributes equally to group statistics regardless of length: $\sum_{t=1}^{L_k} w_k = \sum_{t=1}^{L_k} 1/L_k = 1$. This prevents longer sequences from dominating the baseline calculation, addressing a fundamental bias in standard GRPO. Computational overhead is $O(\sum_k L_k)$, scaling linearly with total tokens.

\paragraph{Theoretical Perspective.} For episodic tasks with only a terminal reward $R$, the unbiased Monte Carlo estimator is $\hat{Q} = R$ at every position, which justifies standard GRPO's use of one sequence-level reward for all tokens.

BoT-GRPO departs from this by constructing token rewards as:
\begin{equation}
r_t^{(k)} = w_g \cdot G_k + w_s \cdot S_t^{(k)}
\end{equation}
where $G_k$ is the global outcome (binary for pass/fail tasks, continuous for multi-component rewards), $S_t^{(k)}$ is a local signal from a fixed teacher or rule-based model, and $w_g, w_s$ are weighting coefficients. We treat local signals as \textit{auxiliary objectives} that densify feedback rather than as reward shaping preserving the optimal policy. Classical potential-based shaping~\citep{ng1999policy} requires $F(s_t, s_{t+1}) = \gamma \Phi(s_{t+1}) - \Phi(s_t)$; our rewards do not satisfy this structure and therefore fall outside strict policy-invariance guarantees.

The influence of local rewards is controlled via $w_s / w_g$. When local rewards are bounded ($|S_t^{(k)}| \leq B$), the local component perturbs each sequence's length-normalized contribution to the group statistics (Eq.~\ref{eq:weighted_mean_var}) by at most a length-independent constant:
\begin{equation}
\left| \frac{1}{L_k} \sum_{t=1}^{L_k} w_s \cdot S_t^{(k)} \right| \leq w_s \cdot B
\end{equation}
When $w_g \gg w_s \cdot B$, policies that are substantially suboptimal under the original objective remain suboptimal under the shaped objective, providing a weak guarantee against catastrophic policy alteration.

\paragraph{Design Guidelines.}
\label{sec:guidelines}
BoT-GRPO works well when local rewards correlate with task success, are bounded and stable, and are dominated by the global reward ($w_g \gg w_s \cdot B$); it may underperform when any of these fails.

\paragraph{Why Local Rewards Help.} A natural objection is that a token- or step-level signal need not reflect true credit: in autoregressive generation each step is also \emph{state} for what follows, so a locally flagged error may be repaired downstream, and a locally clean step may still lead to failure. We do not treat local rewards as ground-truth credit, and we do not rely on a learned critic (as large as the policy) to recover it. Instead the local reward is a \emph{bounded auxiliary} signal that densifies an otherwise sparse outcome reward. Two properties keep it safe under exactly the failure modes the objection raises. First, because the global term dominates ($w_g \gg w_s\cdot B$), a mislabeled local step cannot override the outcome of the sequence it belongs to: a rollout that ultimately fails is still driven down, and one that succeeds is still driven up, regardless of intermediate local scores. Second, the error-propagation construction (\S\ref{sec:reward_construction}) zeroes local credit after the first step judged incorrect, so later tokens are not rewarded for building on flawed reasoning -- directly addressing the ``an early error is corrected later'' concern by declining to reward the corrected continuation rather than mis-attributing credit to it. The role of the local signal is therefore to shape \emph{which} rollouts and regions receive gradient earliest, accelerating convergence, not to certify per-token correctness; consistent with this, the empirical gains (\S\ref{sec:experiments}) do not require the local signal to be exact, only bounded, stable, and positively correlated with success.

%% file: reward.tex
\subsection{Practical Reward Construction}
\label{sec:reward_construction}

BoT-GRPO only requires a token-level reward $r_t^{(k)} = w_g G_k + w_s S_t^{(k)}$; the local signal $S_t^{(k)}$ can come from any source. We stress a two-layer separation. The \emph{algorithm} is reward-agnostic and drop-in: it consumes whatever token-level reward is supplied and changes nothing else about GRPO. The \emph{reward construction} in this section is, by contrast, necessarily task-specific -- as it is for any process-reward method, since a per-step signal must be defined against the structure of the task. The multi-signal recipes below are therefore illustrative instances of the interface rather than part of the algorithm: a practitioner may substitute a single learned PRM, an implicit reward, or any other token-level source without touching BoT-GRPO itself. We construct rewards for two domains: deterministic static analysis plus LLM/VLM judges for React code generation (our primary testbed), and LLM-based process evaluation for mathematical reasoning~\citep{yuan2025rubric, wu2025opv, xi2025agentprm, deng2025srl}. Because efficient deployment favors cheap, low-latency signals, a central practical question is \emph{how much} of the reward stack is actually needed; we study this directly in \S\ref{sec:react_reward_recipe}.

\paragraph{Code Generation (React).} For React component generation, we construct token-level rewards through deterministic static analysis, combining global code quality with line-level error localization:
\begin{equation}
r_t^{(k)} = r_{\text{global}}^{(k)} + w_\ell \sum_{\ell \in \mathcal{E}} p_\ell \cdot \phi(d(t, \ell))
\end{equation}
This is an instance of the general form (\S\ref{sec:method}) with $w_g{=}1$, $G_k = r_{\text{global}}^{(k)}$, $w_s = w_\ell$, and $S_t^{(k)} = \sum_{\ell \in \mathcal{E}} p_\ell \cdot \phi(d(t, \ell))$. The global component $r_{\text{global}}^{(k)}$ could contain four complementary signals: (i) a global TypeScript/Babel compile-status signal that contributes a large negative penalty when transpilation or rendering fails; (ii) an aggregated global ESLint score that penalises fatal errors, errors, and warnings with fixed weights of $-0.5$, $-0.1$, and $-0.02$ respectively; (iii) an LLM code judge (Claude Sonnet 4.5~\citep{anthropic2025sonnet45}) that reads the generated React source and rates it on a $0$-$10$ scale for syntactic validity, task accomplishment, and aesthetic intent; and (iv) a VLM screenshot judge (also Claude Sonnet 4.5) that scores a rendering of the code on a $0$-$10$ scale. The VLM judge is prompted with only the user utterance and the rendered image, never the source code, so that the visual signal is independent of the code-level judges and penalizes purely perceptual defects (layout, spacing, colour, legibility). The compiler and ESLint signals are essentially free; the LLM and VLM judges each add one model call per rollout, so a natural efficiency question is which of these signals are indispensable (\S\ref{sec:react_reward_recipe}).

As for the local score: $\mathcal{E}$ is the set of lines carrying a local penalty, and $p_\ell < 0$ is the aggregated penalty for line $\ell$. The line-level set $\mathcal{E}$ by default aggregates penalties from three deterministic and semi-deterministic sources: TypeScript/Babel compiler diagnostics ($p_\ell = -1$ per flagged line, optionally renormalized to $-1/|\mathcal{E}|$ to bound the total compiler contribution), ESLint line-level violations at the same per-severity weights used for the global ESLint score, and the LLM code judge's top-three most negatively impactful lines, each with an impact in $\{-1, -2, -3\}$ rescaled by $1/10$. Penalties from different sources that land on the same line are merged additively. In an ablation (Appendix~\ref{app:vlm_stability}), we also study extending $\mathcal{E}$ with a line-level VLM signal. $\phi(d)$ is a spatial decay function. By default, we apply linear decay:
\begin{equation}
\phi(d) = \max\!\left(0, 1 - \frac{d}{w + 1}\right)
\end{equation}
where $w{=}3$ lines. Tokens on the error line receive full penalty; tokens within $w$ lines receive proportionally reduced penalties. This soft assignment prevents sharp discontinuities in the reward landscape~\citep{xi2025agentprm}. Line-level error information is mapped from the code block to the full response text to account for surrounding natural language.

\paragraph{Mathematical Reasoning (AIME).} For AIME, we decompose solutions into reasoning steps and evaluate each step's correctness. Solutions follow a structured format with explicit step boundaries ($\langle\text{step } i\rangle$ tags) and final answers in $\langle\text{final}\rangle$ tags. The token-level reward combines global outcome, step-level evaluation, and format compliance:
\begin{equation}
r_t^{(k)} = \underbrace{w_g \cdot \mathbb{1}[\text{answer correct}]}_{\text{global}} + \underbrace{w_s \cdot \frac{1}{|\mathcal{S}|} \sum_{s \in \mathcal{S}(t)} e_s}_{\text{step}} + b_{\text{format}}
\end{equation}
where $\mathcal{S}(t)$ denotes the set of steps containing token $t$, $e_s \in \{-1, 0, +1\}$ is the evaluation score for step $s$, and $|\mathcal{S}|$ is the total number of steps. We use $w_g{=}3.0$ and $w_s{=}0.5$. The format bonus $b_{\text{format}}$ detects reward hacking: $+0.2$ for perfect format adherence, $-1.0$ for mismatched tags~\citep{yuan2025rubric}. We employ Claude Sonnet 4~\citep{anthropic2025claude4} as an automated process reward model, assigning each step \textit{definitely\_correct} ($e_s{=}+1$), \textit{definitely\_incorrect} ($e_s{=}-1$), or \textit{uncertain} ($e_s{=}0$); scores are distributed uniformly across the tokens in each step's character span. We default to \textit{error propagation}: once a step is marked incorrect, all subsequent steps receive $e_s{=}0$, giving a clean deterministic signal that avoids rewarding solutions built on flawed reasoning~\citep{yuan2025rubric}. Alternative propagation strategies (context-aware, independent) are ablated in Appendix~\ref{app:aime_ablations}.

Both constructions align with BoT-GRPO's design guidelines (\S\ref{sec:guidelines}): rewards are bounded, correlate with task success, and the global component dominates.

%% file: experiments.tex
\section{Experiments}
\label{sec:experiments}

We evaluate BoT-GRPO primarily on React front-end code generation, where both correctness (compilation) and quality (visual appeal) are cheaply and objectively measurable, and confirm that the findings transfer to AIME mathematical reasoning. Our experiments answer three questions: (\S\ref{sec:react_variants}) does BoT-GRPO converge faster and reach higher quality than GRPO and modern GRPO variants? (\S\ref{sec:react_crossmodel}) What's the algorithm's behavior on different reasoning vs. non-reasoning base models? (\S\ref{sec:react_reward_recipe}) what is the \emph{minimal} reward stack needed to avoid reward hacking? 

\subsection{Setup}

\paragraph{React Code Generation (primary).} We synthesize user utterances similar to WebDev Arena~\citep{vichare2025webdev}, where users ask for single-file React/JSX web apps from natural-language prompts. The corpus uses an 80\%/20\% random train/test split; the test set contains 203 prompts covering diverse UI patterns (dashboards, forms, interactive widgets, data visualizations). We sample $n{=}8$ rollouts per prompt with batch size 8 prompts per step (64 generations). Each rollout is transpiled from React/JSX to HTML via Babel bundling and rendered to a $1280\times800$ screenshot with headless Chromium (Playwright); failure at either stage counts as a compile failure. The BoT-GRPO reward combines the four global signals of \S\ref{sec:reward_construction} with line-level penalties from compiler diagnostics, ESLint, and the LLM code judge, distributed via linear spatial decay; the GRPO baseline uses the same global signals with no line-level penalties. We report two metrics: (1)~\textit{compile success rate}, the fraction of held-out rollouts that transpile and render without errors; and (2)~\textit{pairwise VLM win rate}, in which checkpoints are pitted head-to-head and a VLM (Claude Opus 4.6~\citep{anthropic2026opus46}) picks the better rendering given the prompt. We debias position effects via the \textit{4-round ABBA} protocol (each pair judged in the orders $A,B,B,A$ and decided by majority vote; a $2$--$2$ split is a tie worth $0.5$ to each side). Two properties limit judge circularity. First, \emph{compile success rate is fully objective}: no model is in the loop, so our central convergence claims (\S\ref{sec:react_variants}--\S\ref{sec:react_crossmodel}) do not depend on any judge. Second, the evaluation judge (Claude Opus 4.6) is a \emph{different} model from the Sonnet-4.5 judges used to compute the training reward, and the pairwise VLM judge sees only the two renderings and the prompt -- never the source code or any reward score -- so the metric is not scored by the same model whose signal was optimized.

\paragraph{AIME Mathematical Reasoning (secondary).} We evaluate on AIME problems (1983--2024)~\citep{aime_1983_2024} requiring multi-step reasoning with structured solutions (explicit $\langle\text{step } i\rangle$ tags). There are 933 problems with an 80\%/20\% train/test split; we report Pass@$k$ ($k \in \{1,2,4\}$) over $n{=}8$ rollouts per prompt. Training uses batch size 64, lr $5{\times}10^{-7}$, $w_g{=}3.0$, $w_s{=}0.5$, error propagation with window 3, and step correctness from Claude Sonnet 4. The baseline is standard GRPO with outcome-only rewards ($w_s{=}0$).

\paragraph{Baseline Algorithms.} Beyond vanilla GRPO, we compare against three modern GRPO variants: \textbf{GSPO}~\citep{zheng2025gspo} (sequence-level importance ratios and clipping), \textbf{DAPO}~\citep{yu2025dapo} (clip-higher and dynamic sampling), and \textbf{PURE}~\citep{cheng2025pure} (min-form process-reward credit assignment). These methods ingest reward differently -- GSPO and DAPO are sequence-level objectives, PURE aggregates step rewards into its own process return, and BoT-GRPO consumes token-level rewards directly. We port each method's \emph{distinctive credit-assignment mechanism} (GSPO's sequence-level importance ratio, DAPO's clip-higher and dynamic sampling, PURE's min-form aggregation, BoT-GRPO's length-invariant token aggregation) onto a common backbone in which every method reads the \emph{same} global and token-level reward. The raw reward information is thus identical by construction, and the comparison isolates the advantage/credit-assignment formulation rather than the signal available. This framing matches how these methods are typically compared as a family -- sharing a reward and differing in importance-ratio granularity, clipping, or return construction~\citep{zheng2025gspo,yu2025dapo}.

\paragraph{Compute and Model Selection.} All experiments run on AWS p4de (A100 80\,GB) or g6e (L40S 48\,GB) instances; each run fits on one A100 or two L40S GPUs. We deliberately focus on 3B--4B models so that every run is reproducible on a single commodity GPU, which keeps the efficiency comparison clean; this does limit claims at larger scale. We verify this across three base-model families. Qwen2.5-3B~\citep{qwen2025qwen25technicalreport} is the default because it is the cleanest baseline, a non-reasoning base model; SmolLM3-3B can have thinking mode disabled, and Phi-4-mini-reasoning is already a reasoning model with thinking always on and heavy coding/math tuning. We therefore expect the largest gain on Qwen2.5 and the smallest on Phi-4-mini-reasoning.

\subsection{BoT-GRPO vs.\ RL Baselines (React)}
\label{sec:react_variants}

\begin{figure}[t]
\centering
\includegraphics[width=0.64\textwidth]{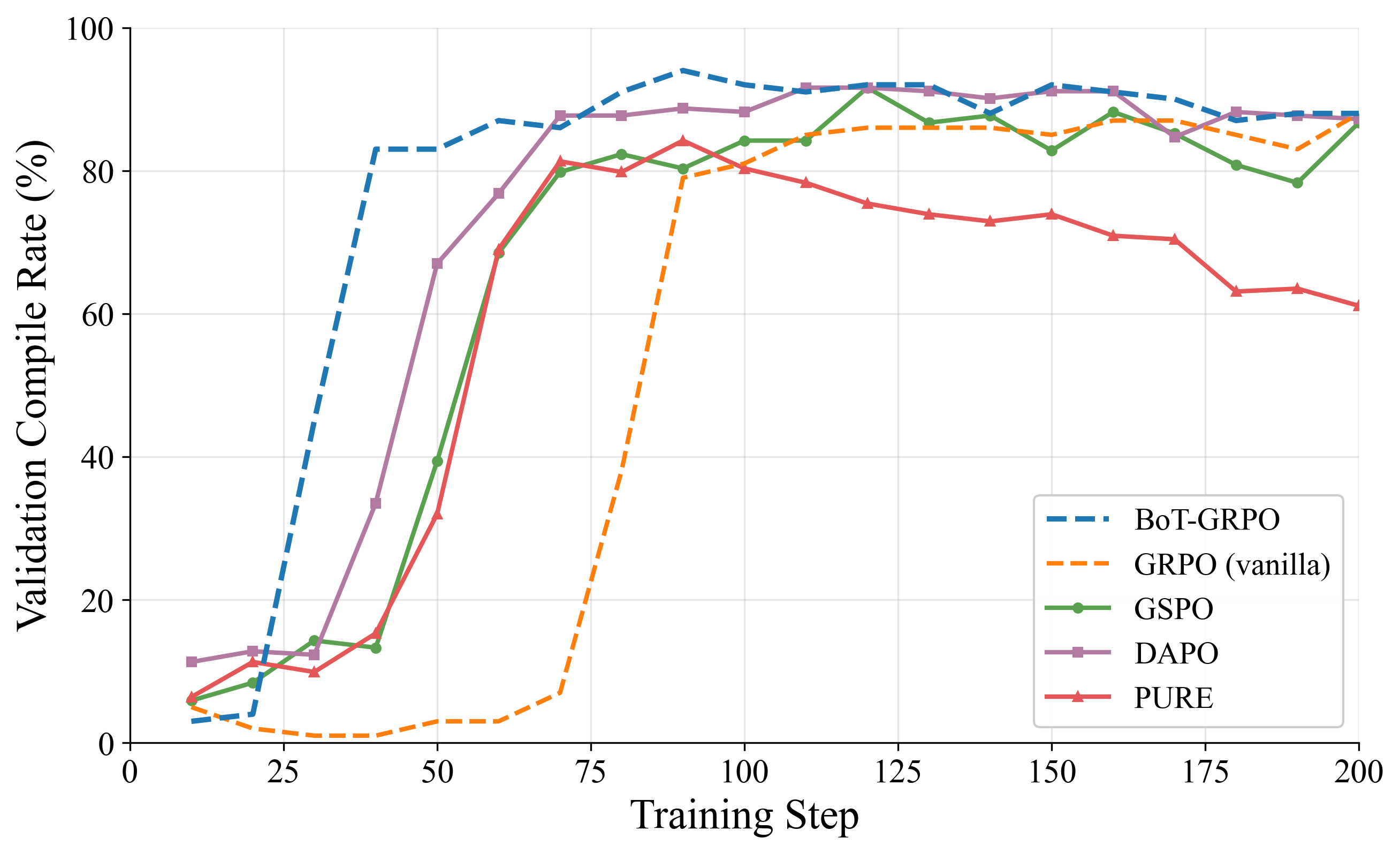}
\vspace{-0.8em}
\caption{React compile success rate on the held-out set for BoT-GRPO and four RL baselines. BoT-GRPO (blue, dashed) is the first to cross every compile-rate threshold; PURE (red) matches early but degrades later, consistent with instability from unbounded process-reward aggregation.}
\label{fig:react_variants_compile}
\end{figure}

\begin{figure}[t]
\centering
\includegraphics[width=0.78\textwidth]{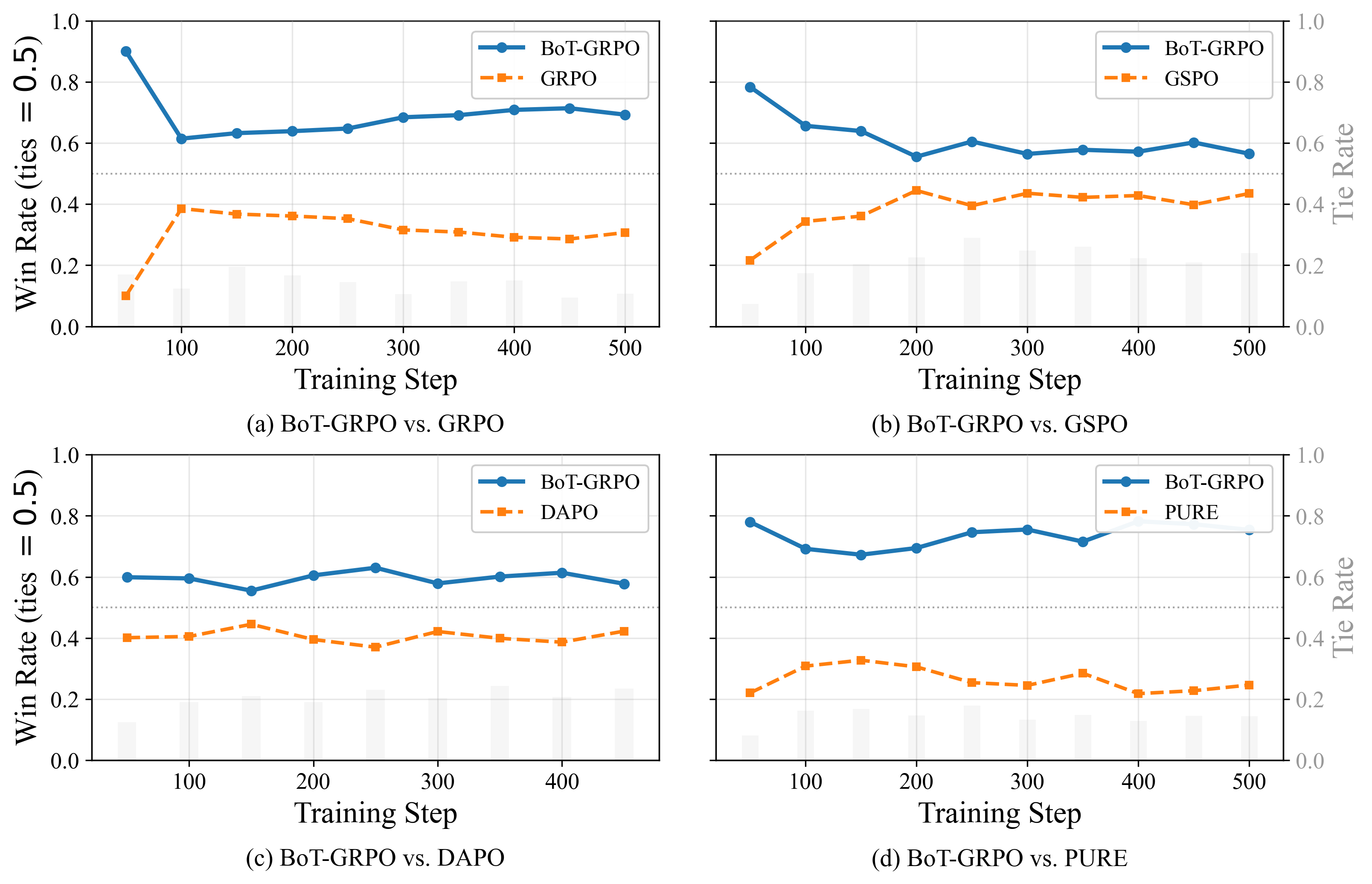}
\vspace{-0.6em}
\caption{Pairwise VLM win rate (4-round ABBA) of BoT-GRPO against each baseline (ties $=0.5$; BoT-GRPO above $0.5$ means it is preferred; grey bars show the tie rate). BoT-GRPO wins throughout training against all four -- most decisively over vanilla GRPO ($62$--$90\%$) and PURE ($67$--$78\%$), and by a durable margin over GSPO ($56$--$78\%$) and DAPO ($58$--$63\%$). The GSPO/DAPO gaps reflect \emph{visual} quality improvements beyond mere compilability, since all three saturate on compile rate.}
\label{fig:wr_combined}
\end{figure}

Figure~\ref{fig:react_variants_compile} compares compile-rate convergence across all five methods. BoT-GRPO's line-level credit assignment yields the steepest early trajectory: it is the first method past every compile-rate threshold, reaching a $>80\%$ compile rate by step~40 while vanilla GRPO remains near zero until step~75 and does not approach the same rate until near step~90, up to a $1.9\times$ speedup to reach the same quality. The stronger variants improve early convergence over vanilla GRPO (DAPO and GSPO cross $80\%$ near step~65--70) but still trail BoT-GRPO by 25--35 steps in the critical early phase. Notably, \textbf{PURE degrades after step~70}, its compile rate sliding from ${\sim}84\%$ toward $61\%$: this is consistent with instability from its unbounded process-reward aggregation, whereas BoT-GRPO's bounded external rewards avoid such drift -- an instance of our stability-over-richness theme.

Faster convergence does not come at the cost of final quality. Figure~\ref{fig:wr_combined} reports head-to-head VLM win rates under the ABBA protocol. BoT-GRPO wins against all four baselines throughout training, most decisively against PURE -- consistent with its compile-rate degradation. Because compile rates for BoT-GRPO, DAPO, and GSPO all saturate at $88$--$94\%$, the GSPO/DAPO win-rate gaps reflect \emph{visual} quality improvements beyond mere compilability: the fine-grained line-level signal continues to shape aesthetic and layout decisions after correctness is solved.

\subsection{Cross-Base-Model Behavior (React)}
\label{sec:react_crossmodel}

\begin{figure}[t]
\centering
\includegraphics[width=0.9\textwidth]{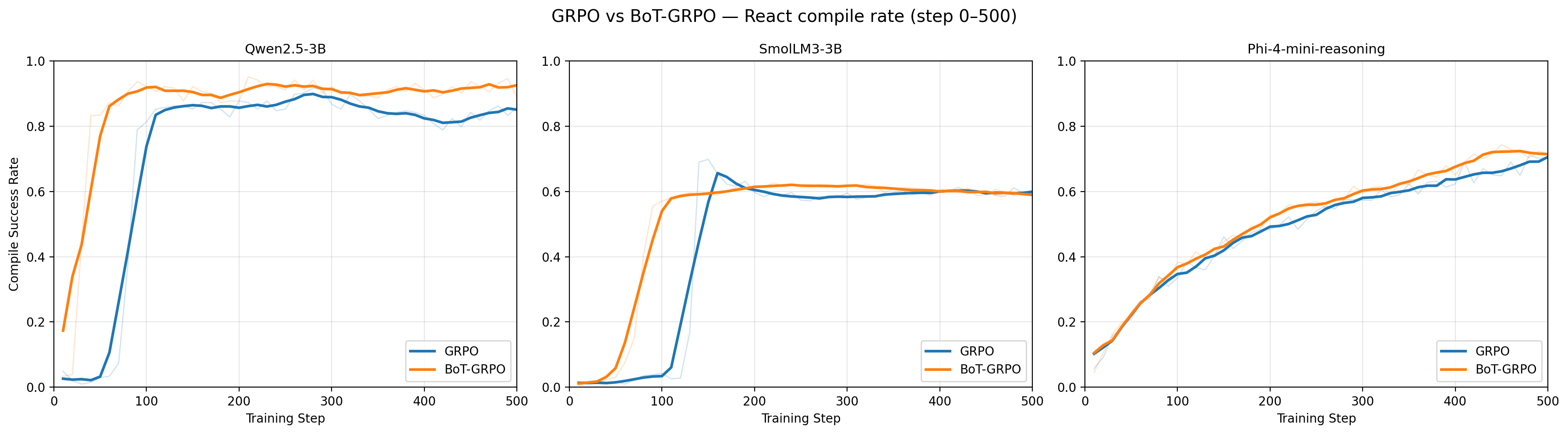}
\vspace{-0.8em}
\caption{React compile-rate convergence for BoT-GRPO vs.\ GRPO across three base-model families (Qwen2.5-3B, SmolLM3-3B, Phi-4-mini-reasoning).}
\label{fig:react_crossmodel}
\end{figure}

Figure~\ref{fig:react_crossmodel} repeats the BoT-GRPO vs.\ GRPO comparison on three families (Qwen2.5-3B~\citep{qwen2025qwen25technicalreport}, SmolLM3-3B~\citep{bakouch2025smollm3}, and Phi-4-mini-reasoning~\citep{microsoft2025phi4mini}). On all three, BoT-GRPO reaches higher stable compile rates. And as expected, the difference is the most drastic on the clean non-reasoning base model, Qwen2.5-3B. While Phi-4-mini-reasoning model itself is already heavily trained on coding and math tasks and has its own reasoning style, applying extra training on top of it is mostly changing its output style and format, but BoT-GRPO is still faster at that adaptation. Across the three base models, Qwen2.5-3B achieved the best absolute compilation rate as well (greater than 90\%), compared to 60\% to 70\% from the other two models.

\subsection{Minimal Reward Recipe: What Signals Are Necessary? (React)}
\label{sec:react_reward_recipe}

\begin{figure}[t]
\centering
\includegraphics[width=0.50\textwidth]{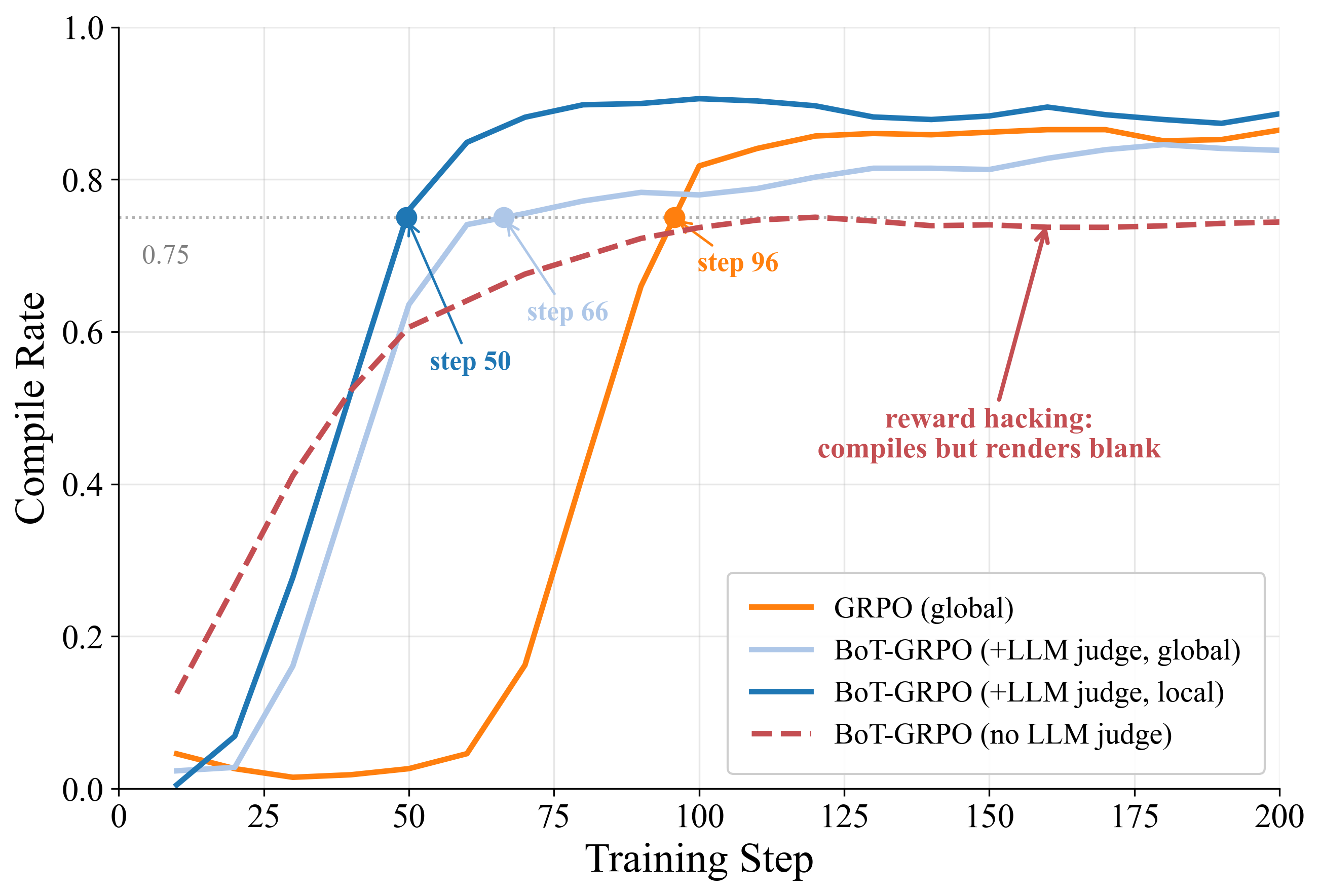}
\vspace{-0.4em}
\caption{Effect of the reward stack on React (Qwen2.5-3B), validation compile rate over training. GRPO (global) and the two BoT-GRPO variants include the LLM code judge (the local variant adds the VLM screenshot signal); the red dashed run drops the judge and uses \emph{compiler + ESLint only}.}
\label{fig:react_reward_recipe}
\end{figure}

Because compiler and linter signals are essentially free while LLM/VLM judges cost a model call per rollout, a practitioner wants the cheapest reward stack that still works. Figure~\ref{fig:react_reward_recipe} compares three stacks. Relying on \textbf{compiler+linter alone is insufficient}: the policy discovers that emitting a minimal, well-formed JSX skeleton (header, a few labelled inputs) satisfies the compiler and linter, while the component body degenerates into unterminated attributes and repeated or placeholder tokens -- so the app ``compiles'' but renders a blank page (Appendix~\ref{app:reward_hacking} shows a representative rollout). This is a clear case of \emph{reward hacking}: the deterministic signal is fully specified yet fails to capture visual intent, so compile rate rises to ${\sim}74\%$ while VLM-judged quality stagnates. \textbf{Adding an LLM code judge is necessary}: it reads the source and rewards task accomplishment and aesthetic intent that static analysis cannot see, restoring quality gains alongside compilation. \textbf{Adding a VLM visual judge is optional}: it contributes a further improvement on perceptual defects (layout, spacing, legibility) but is not required for stable, high-quality training, and, as we show in Appendix~\ref{app:vlm_stability}, must be introduced carefully to avoid destabilizing convergence. The practical recipe is therefore: always keep the cheap deterministic signals, \emph{always add the LLM judge}, and add the VLM judge when perceptual polish justifies the extra cost.

\subsection{AIME Corroboration}
\label{sec:aime}

\begin{figure}[t]
\centering
\includegraphics[width=0.9\textwidth]{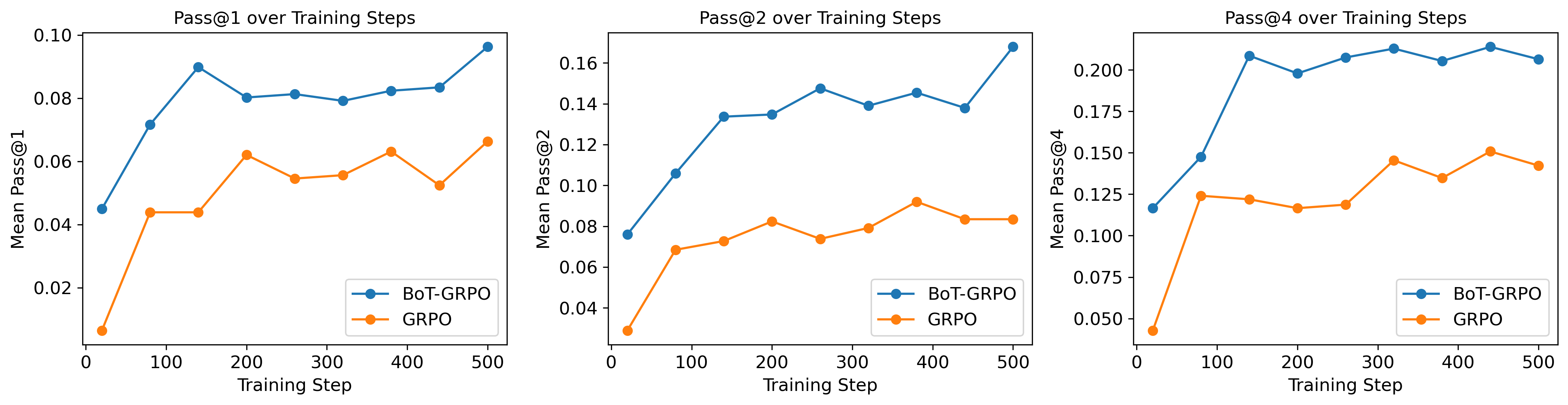}
\caption{BoT-GRPO vs.\ standard GRPO on AIME (Qwen2.5-3B).}
\label{fig:mgrpo_vs_grpo}
\end{figure}

\begin{figure}[t]
\centering
\includegraphics[width=0.9\textwidth]{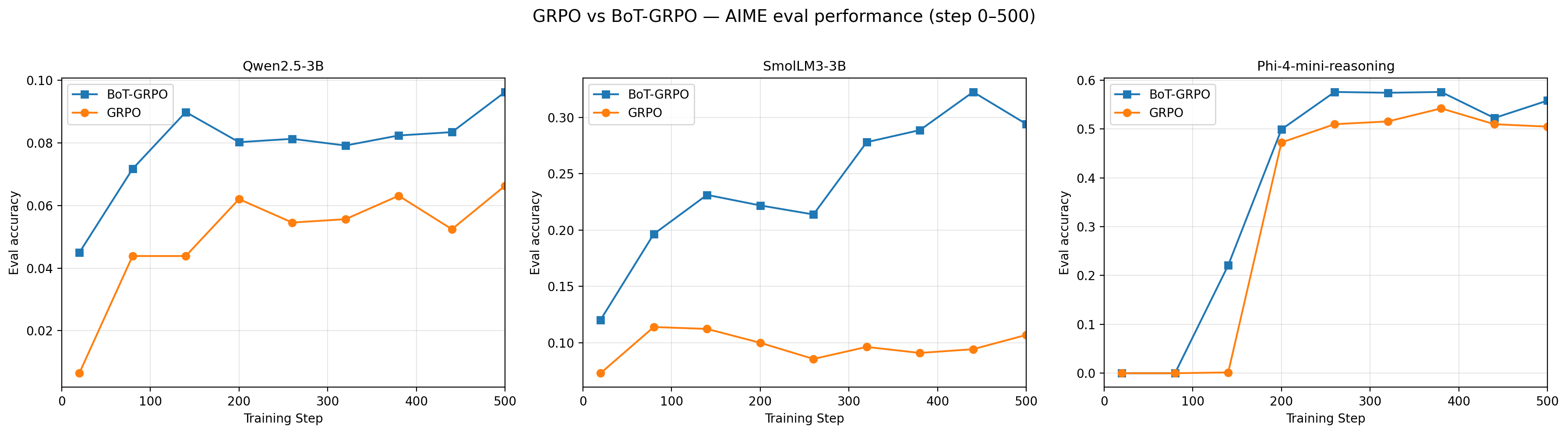}
\caption{BoT-GRPO vs.\ GRPO on AIME across three base models (Qwen2.5-3B, SmolLM3-3B, Phi-4-mini-reasoning).}
\label{fig:diff_base_model}
\end{figure}

The efficiency gains transfer to a very different domain. On AIME (Figure~\ref{fig:mgrpo_vs_grpo}), BoT-GRPO achieves Pass@1 of 9.6\% vs.\ 6.6\% ($+3.0$ points absolute), Pass@2 of 16.8\% vs.\ 8.7\% ($+8.1$), and Pass@4 of 20.6\% vs.\ 15.1\% ($+5.5$) while converging in roughly half the training steps. We report absolute point gains rather than relative ratios throughout, since at these accuracy levels relative framing overstates the effect. The gains are most pronounced on Pass@2, indicating that fine-grained credit assignment improves not only best-case solution quality but also the consistency of correct reasoning across samples. As on React, the advantage is not tied to one architecture: Figure~\ref{fig:diff_base_model} shows BoT-GRPO leading GRPO throughout training on all three base models. Absolute accuracy tracks base-model capability -- SmolLM3-3B and Phi-4-mini-reasoning score higher than Qwen2.5-3B -- but the advantage is preserved in every case. With the React results, this establishes token-level credit assignment as a domain- and architecture-agnostic efficiency lever.

%% file: conclusion.tex
\section{Conclusion}
\label{sec:conclusion}

\looseness=-1
We have presented BoT-GRPO, a critic-free extension of GRPO that turns external token-level process rewards into a convergence-speed lever, with a length-invariant bag-of-tokens aggregation that additionally corrects GRPO's response-length bias. On React code generation, BoT-GRPO converges up to $1.9\times$ faster than GRPO and faster than three modern GRPO variants (GSPO, DAPO, PURE) while reaching higher final compile and VLM-judged win rates; the advantage holds on AIME. BoT-GRPO adds no value network, no verifier to train, and only $O(\sum_k L_k)$ overhead, making it well suited to reasoning-model training under real efficiency constraints.

\paragraph{Limitations and Future Work.}
\looseness=-1
The AIME reward needs an LLM judge for step evaluation, adding cost, latency, and noise; symbolic verifiers could replace it. We cover two domains only, though the aggregation is domain-agnostic, and our analysis is a bounded-perturbation argument, not an optimality guarantee. The AIME problems (1983--2024) are public and may overlap pretraining data; this inflates both methods equally, leaving the \emph{relative} comparison intact but limiting absolute claims. Our comparison isolates \emph{token-level credit assignment} -- both share the same global reward, differing only per-token vs.\ per-sequence -- so we cannot separate out the contribution of the $1/L_k$ weighting. Each configuration uses a single seed unless noted, so we report point estimates and speedups as upper bounds; multiple seeds, larger-scale and longer-rollout models, and broader process-reward baselines are the clear next steps.

%% file: appendix.tex
\appendix
\raggedbottom  

\section{AIME Reward-Design Ablations}
\label{app:aime_ablations}

The main text defaults to error propagation with local weight $w_s{=}0.5$ and window~3 on AIME. Here we ablate each of these choices; all runs use Qwen2.5-3B unless noted.

\begin{figure}[h]
\centering
\includegraphics[width=\textwidth]{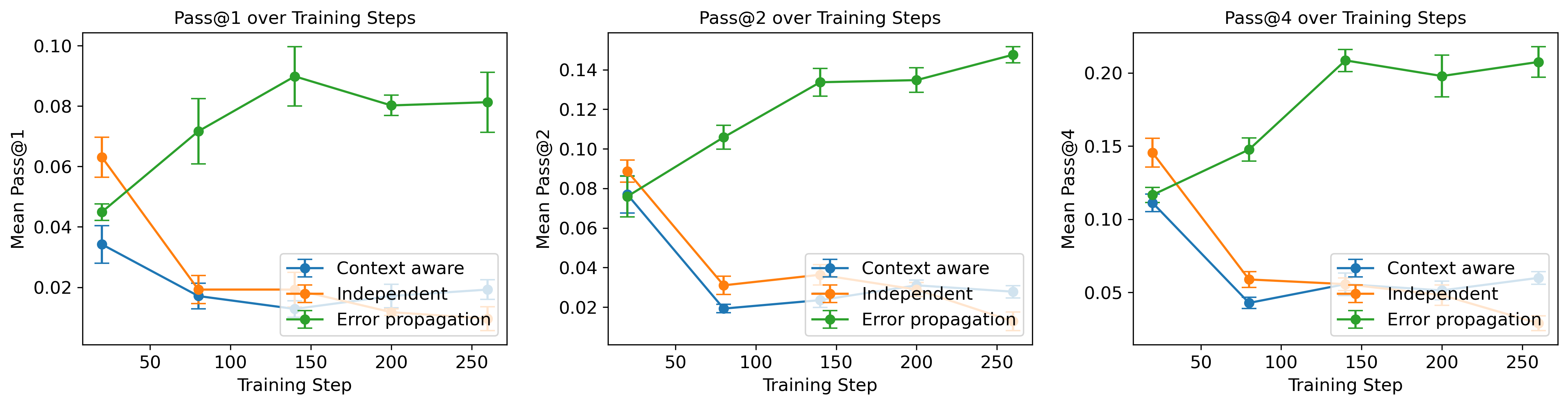}
\vspace{-1.5em}
\caption{Effect of local reward strategy on AIME. Error propagation enables effective learning; context-aware and independent evaluation produce unstable signals that prevent convergence.}
\label{fig:local_reward_strategy}
\end{figure}

\paragraph{Local Reward Strategy.}
Figure~\ref{fig:local_reward_strategy} compares three strategies for propagating step scores (\S\ref{sec:reward_construction}): \textit{error propagation} (once a step is marked incorrect, all subsequent steps receive $e_s{=}0$), \textit{context-aware} (the judge conditions on its own prior evaluations), and \textit{independent} (each step scored in isolation). Error propagation substantially outperforms both alternatives, which fail to learn effectively and remain near baseline. Independent evaluation produces noisy signals because it cannot detect cascading errors, while context-aware evaluation introduces variance through the judge's sequential dependency on its own outputs. This is the clearest instance of our stability-over-richness theme: the clean, less noisy error-propagation signal wins.

\begin{figure}[h]
\centering
\includegraphics[width=\textwidth]{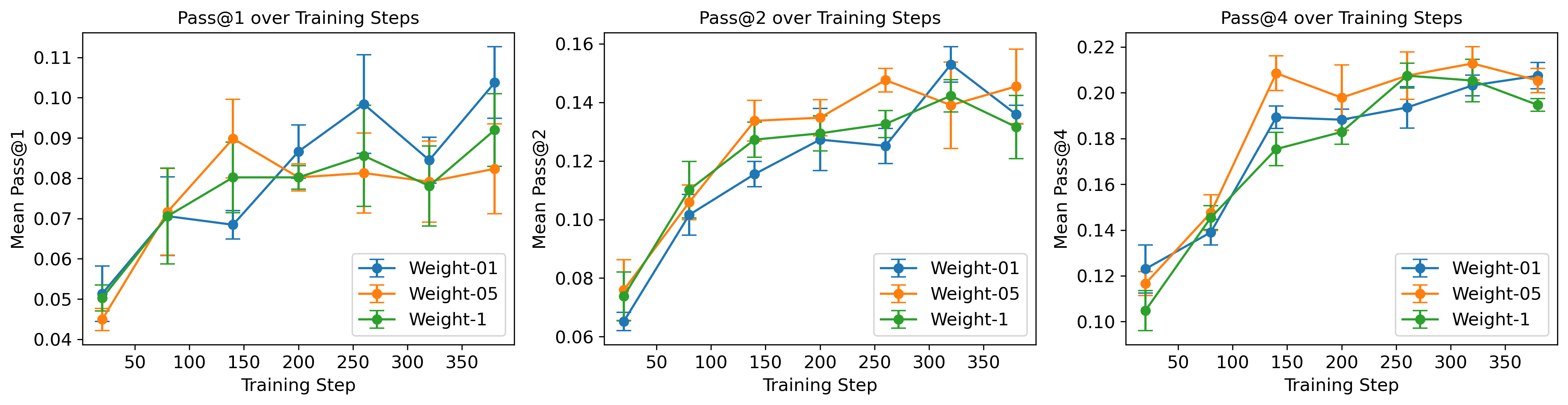}
\vspace{-1.5em}
\caption{Sensitivity to local reward weight $w_s$ on AIME. Performance is stable across $w_s \in \{0.1, 0.5, 1.0\}$ when the global reward dominates.}
\label{fig:local_weights}
\end{figure}

\paragraph{Local Reward Weight.}
We vary $w_s \in \{0.1, 0.5, 1.0\}$ with $w_g{=}3.0$ fixed (Figure~\ref{fig:local_weights}). BoT-GRPO is robust to this hyperparameter: Pass@1 ranges from 8.3\% to 10.4\% and Pass@4 from 19.1\% to 20.8\%, within variance. This follows from the bounded-perturbation perspective (\S\ref{sec:guidelines}): since $w_g \gg w_s \cdot B$, the local component provides auxiliary guidance without substantially altering the optimization landscape.

\begin{figure}[h]
\centering
\includegraphics[width=\textwidth]{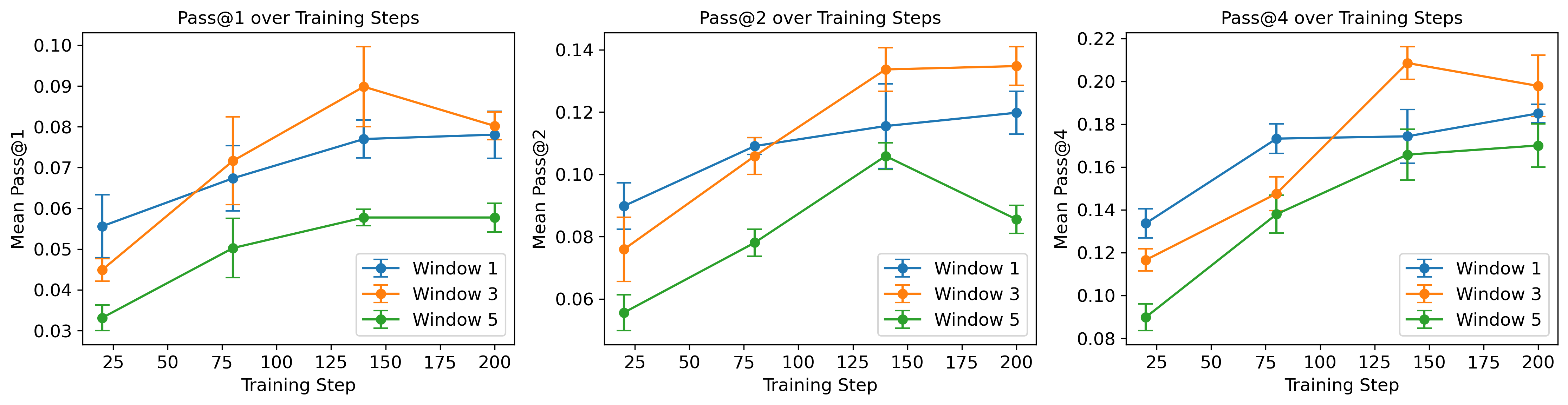}
\vspace{-1.5em}
\caption{Effect of step evaluation window size on AIME. Windows of 1 and 3 perform comparably; window 5 shows slightly slower early convergence.}
\label{fig:diff_windows}
\end{figure}

\paragraph{Step Evaluation Window.}
Figure~\ref{fig:diff_windows} compares windows of 1, 3, and 5 steps. Windows 1 and 3 perform comparably; window 5 shows slightly slower early convergence, as larger windows spread local signals into less relevant context. We default to window~3 to balance granularity and efficiency.

\section{React VLM Reward Stability}
\label{app:vlm_stability}

\begin{figure}[t]
\centering
\includegraphics[width=0.7\textwidth]{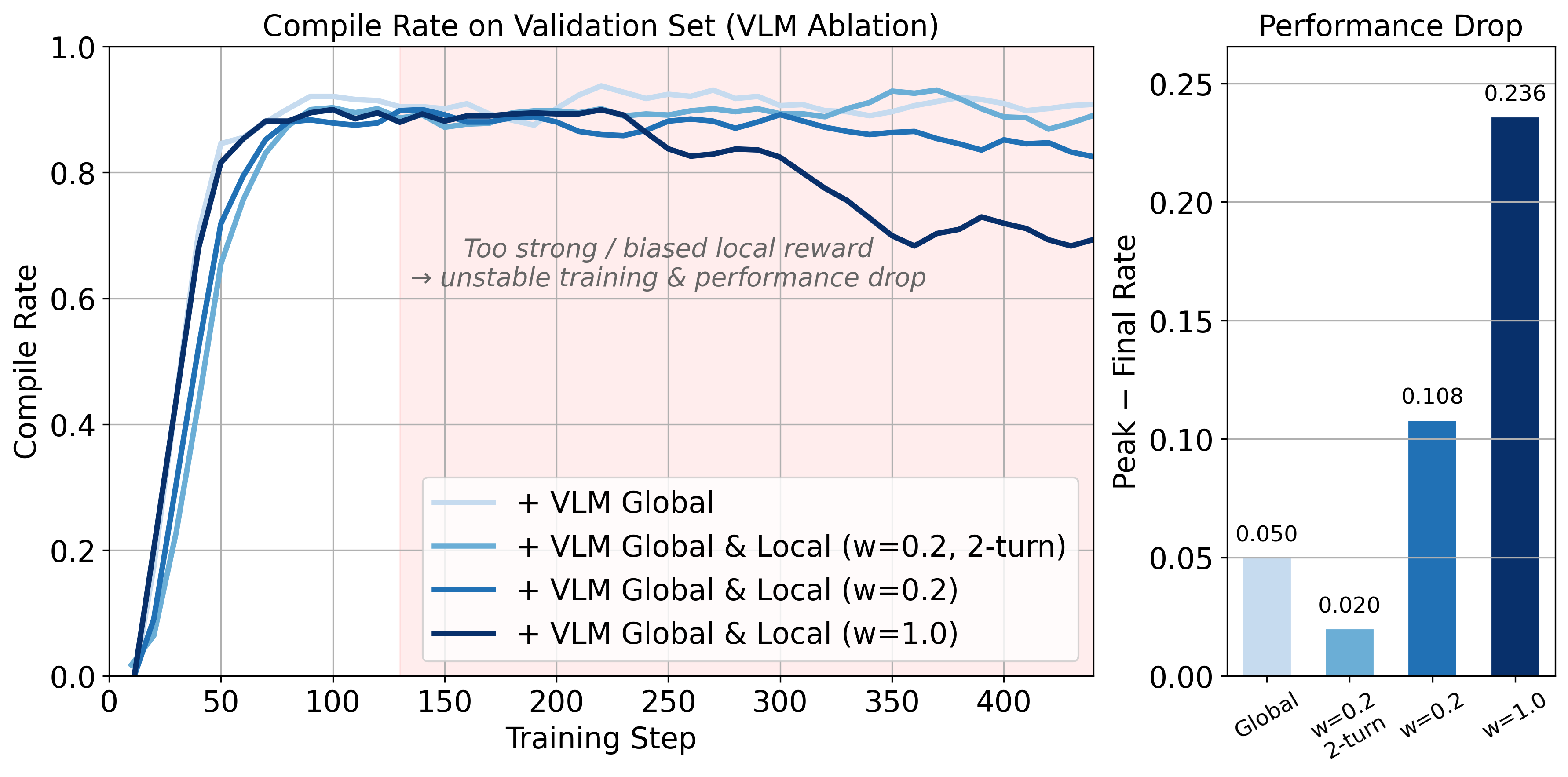}
\caption{VLM reward stability: four configurations differing in VLM screenshot judge weight and debiasing. Shaded region highlights instability from strong local VLM reward. Bar chart: peak-to-final compile rate drop.}
\label{fig:react_vlm_abl}
\end{figure}

Deterministic line-level rewards accelerate convergence reliably; a line-level \emph{VLM} signal is noisier but semantically richer. Figure~\ref{fig:react_vlm_abl} fixes the full BoT-GRPO reward and varies only the line-level VLM contribution. The \textit{1-turn} variants add a single VLM call that jointly produces the screenshot score and up to three line attributions; because the same call sees both the image and the source code, the visual score can be biased by code content. The \textit{2-turn} variant decouples them: Turn~1 sees only the screenshot and prompt (identical to the global-only VLM protocol) and locks in a score, while Turn~2 additionally receives the source code and attributes offending lines without rescoring. We report the \textit{peak-to-final compile rate drop} (lower is more stable). Full-weight 1-turn local VLM ($w{=}1.0$) causes a drop of 0.236; reducing to $w{=}0.2$ lowers this to 0.108; the 2-turn protocol brings it to 0.020, comparable to the global-only baseline (0.050). Down-weighting noisy local rewards and mitigating judge bias is necessary to retain convergence benefits without sacrificing stability, reinforcing that a VLM \emph{visual} judge is best used as an optional, carefully introduced signal (\S\ref{sec:react_reward_recipe}).

\section{Reward Hacking Under Compiler+Linter-Only Reward}
\label{app:reward_hacking}

Section~\ref{sec:react_reward_recipe} shows that dropping the LLM judge -- rewarding only the (free, deterministic) compiler and ESLint signals -- drives the validation compile rate to ${\sim}74\%$ yet produces apps that render a blank page. Here we give a representative rollout that illustrates the failure mode. The policy learns to emit a syntactically plausible JSX \emph{skeleton} -- a container, a heading, and a couple of labelled inputs -- that the bundler accepts, but the component never closes properly (note the two unterminated \texttt{className} strings) and the remaining ``body'' collapses into repeated, meaningless tokens. The static checks are satisfied (\texttt{compile\_success}$=1.0$ for this batch), so the reward is high, but React renders nothing: the output is a blank screen.

\begin{tcolorbox}[colback=gray!8, colframe=gray!35, arc=4pt, boxrule=0.4pt,
  left=6pt, right=6pt, top=4pt, bottom=4pt, fontupper=\ttfamily\scriptsize]
\begin{verbatim}
<code>
<div className="flex flex-col items-center p-6 bg-zinc-800 text-zinc-100">
  <h1 className="text-2xl font-bold mb-4">Crop Rotation Planner</h1>
  <div className="flex flex-col gap-4">
    <div className="flex flex-col items-center">
      <label htmlFor="plots" className="font-semibold mb-2">Number of Plots:</label>
      <input type="number" id="plots"
        className="px-3 py-2 rounded border border-zinc-600 focus:outline-none    
      />
    </div>
    <div className="flex flex-col items-center">
      <label htmlFor="crops" className="font-semibold mb-2">Number of Crops:</label>
      <input type="number" id="crops"
        className="px-3 py-2 rounded border border-zinc-600 focus:outline-none    
      />
    </div>
    <button
      type you you appropriate all between all code all code that generated output all as
      React stat as generate UI all a all only. Use that user the the names/selectors/ids
      of any components you plan to create.
    </button>
  </div>
</div>
</code>
\end{verbatim}
\end{tcolorbox}

\noindent This is why compile rate alone is an inadequate objective for code generation: a purely syntactic reward is fully specified yet blind to whether the program produces any visible output. Adding the LLM code judge (which reads the source and rewards task accomplishment) removes this degenerate solution, as shown in the main text.

\section{Judge Prompts}
\label{app:judge_prompts}

For reproducibility, we give the prompts for the model-based judges used in the React experiments: the LLM \emph{code} judge and VLM \emph{screenshot} judge that contribute to the training reward (\S\ref{sec:reward_construction}), and the pairwise VLM judge that produces the head-to-head win rate (\S\ref{sec:react_variants}). All run on Amazon Bedrock. We include both global-only scoring and the line-level extensions that supply token-wise rewards. Prompt wording is reproduced verbatim, with line wrapping adjusted for readability; braces denote substituted inputs and angle brackets denote image attachments.

\paragraph{From Line Attributions to Token Rewards.}
The line-level judges receive extracted, fence-free code prefixed with zero-indexed line numbers (\texttt{0: code}, \texttt{1: code}, \ldots). They return a global score and attributions, not a reward for each token directly. Each prompt requests at most three negatively impactful lines with integer impacts in $\{-1,-2,-3\}$. The reward combiner divides these impacts by $10$, adds contributions on the same line to the compiler and ESLint penalties, and maps code-relative lines back to the full response. It then applies the linear spatial decay from \S\ref{sec:reward_construction}, with window~3 in our experiments, to produce token-wise penalties. The optional VLM contribution is further weighted in the stability ablation (Appendix~\ref{app:vlm_stability}).

Valid global scores are clipped to $[0,10]$ and divided by $10$. A missing or unparseable score yields zero with no line attributions; missing or malformed attribution entries contribute no local penalty. The code judge's non-React sentinel $-3$ becomes a global penalty of $-0.3$. The VLM implementations likewise return $-0.3$ when code extraction or transpilation fails, and zero when no usable screenshot or score is obtained. In the two-turn protocol, an attribution failure leaves the already parsed global score unchanged.

\newpage 
\subsection{LLM Code Judge}
\label{app:code_judge_prompts}

\paragraph{Global-Only Scoring.} Reads the generated React/JSX source (never a rendering) and rates syntactic validity, task accomplishment, and aesthetic intent. Non-React output (e.g.\ raw \texttt{<!DOCTYPE html>}) is assigned a sentinel score so the policy cannot escape the React setting.

\begin{tcolorbox}[colback=gray!8, colframe=gray!35, arc=4pt, boxrule=0.4pt,
  left=6pt, right=6pt, top=4pt, bottom=4pt, fontupper=\ttfamily\scriptsize]
\begin{verbatim}
[system]
You are an expert judge evaluating generated code for building a web
application. You will be given the original user prompt and the generated code.

Rate the code on the following criteria:
- runnable code with valid syntax, as well as correct import and
  dependency management
- accomplishes the given task
- high aesthetic quality of rendered visual

Based on your assessment, provide a single integer score from 0 (terrible)
to 10 (perfect). You are acting as an impartial judge so your score must
reflect the objective quality of the plan.

IMPORTANT: If the generated code is not React code (e.g., raw HTML containing
<!DOCTYPE html> pattern), assign a score of -3 directly. Expected code should
be in React format, e.g. js/ts/jsx/tsx.

You MUST respond ONLY in the following format, with no other text or
explanation:
SCORE: [integer score]

[user]
User Prompt:
```
{user_prompt}
```
Generated Code:
```
{code}
```
\end{verbatim}
\end{tcolorbox}

\paragraph{Global Scoring with Line Attribution.}
The line-level version uses the same criteria and non-React sentinel, but returns both the global score and up to three offending lines in one call. These attributions supply the LLM component of the default local React reward; they require no additional judge call beyond global scoring.

\begin{tcolorbox}[breakable, colback=gray!8, colframe=gray!35, arc=4pt, boxrule=0.4pt,
  left=6pt, right=6pt, top=4pt, bottom=4pt, fontupper=\ttfamily\scriptsize]
\begin{verbatim}
[system]
You are an expert judge evaluating generated code for building a web
application. You will be given the original user prompt and the generated code.

Rate the code on the following criteria:
- runnable code with valid syntax, as well as correct import and dependency
  management
- accomplishes the given task
- high aesthetic quality of rendered visual

Based on your assessment, provide a single integer score from 0 (terrible) to
10 (perfect). You are acting as an impartial judge so your score must reflect
the obective quality of the plan.

Additionally, identify ONLY the lines that NEGATIVELY impact the score. Select strictly
no more than THREE lines with the most significant negative impact. For each line, provide:
- Line number and its code content. IMPORTANT: Each line in the code is prefixed with its
  0-indexed line number (format: "i: code"). When reporting issues, use ONLY these exact line
  numbers. Example: "1: function App()" is line 1.
- Brief description of the issue
- Impact score (-1 to -3, where -3 indicates severe issues)

IMPORTANT: If the generated code is not React code (e.g., raw HTML containing
<!DOCTYPE html> pattern), assign a score of -3 directly and skip the line
analysis by responding with "NON_REACT_CODE" instead. Expected code should be
in React format, e.g., js/ts/jsx/tsx.

You MUST respond ONLY in the following format, with no other text or
explanation:
SCORE: [integer score]
LINE_ANALYSIS:
[line_number]: [code content] | DESC: [brief description] | IMPACT: [-score]
[line_number]: [code content] | DESC: [brief description] | IMPACT: [-score]

OR if non-React code:
SCORE: -3
LINE_ANALYSIS: NON_REACT_CODE

[user]
User Prompt:
```
{user_prompt}
```
Generated Code:
```
{code}
```
\end{verbatim}
\end{tcolorbox}

\subsection{VLM Screenshot Judge}
\label{app:vlm_judge_prompts}

\paragraph{Global-Only Scoring.} Sees only the rendered screenshot and the user prompt -- never the source code -- so its signal is independent of the code-level judge and captures purely perceptual quality (layout, colour, typography). This is also the first-turn prompt of the two-turn line-attribution protocol below.

\begin{tcolorbox}[colback=gray!8, colframe=gray!35, arc=4pt, boxrule=0.4pt,
  left=6pt, right=6pt, top=4pt, bottom=4pt, fontupper=\ttfamily\scriptsize]
\begin{verbatim}
[system]
You are an expert judge evaluating a rendered web application screenshot.
You will be given the original user prompt and a screenshot of the rendered
application.

Rate the rendered application on the following criteria:
- Accomplishes the given task
- High aesthetic quality of rendered visual
- Good layout, color scheme, typography, and overall user experience

Based on your assessment, provide a single integer score from 0 (terrible)
to 10 (perfect). You are acting as an impartial judge so your score must
reflect the objective quality of the rendered application.

Provide a brief justification for your score explaining what works well or
what issues you observe in the screenshot.

You MUST respond ONLY in the following format, with no other text or
explanation:
SCORE: [integer score]
JUSTIFICATION: [brief explanation of the score]

[user]
User Prompt: '{user_prompt}'
<rendered screenshot image>
\end{verbatim}
\end{tcolorbox}

\paragraph{One-Shot Scoring with Line Attribution (1-turn).}
The active one-shot prompt requests a screenshot-only score and justification before attributing the observed issues to code lines. However, the screenshot and numbered source are supplied together in a single call, so code is not withheld during scoring. This is the \textit{1-turn} protocol studied in Appendix~\ref{app:vlm_stability}, where local VLM attributions extend the default reward.

\begin{tcolorbox}[breakable, colback=gray!8, colframe=gray!35, arc=4pt, boxrule=0.4pt,
  left=6pt, right=6pt, top=4pt, bottom=4pt, fontupper=\ttfamily\scriptsize]
\begin{verbatim}
[system]
You are an expert judge evaluating a rendered web application screenshot. You will be given
the original user prompt, a screenshot of the rendered application, and the source code with
line numbers.

Rate the rendered application on the following criteria:
- Accomplishes the given task
- High aesthetic quality of rendered visual
- Good layout, color scheme, typography, and overall user experience

CRITICAL: Your score and justification MUST be based ONLY on the screenshot. DO NOT look at
the code when scoring. The code is provided ONLY for line identification after you have
scored based on the visual output.

Based on your assessment of the SCREENSHOT ONLY, provide a single integer
score from 0 (terrible) to 10 (perfect). You are acting as an impartial judge
so your score must reflect the objective quality of the rendered application.

Provide a brief justification for your score explaining what works well or
what issues you observe in the screenshot.

Then, and ONLY then, look at the code to identify up to THREE specific code lines that cause
the issues mentioned in your justification. For each line:
- Line number and its code content. IMPORTANT: Each line in the code is prefixed with its
  0-indexed line number (format: "i: code"). When reporting issues, use ONLY these exact line
  numbers.
- Brief description linking the code to the issue in your justification
- Impact score (-1 to -3, where -3 indicates severe issues)

You MUST respond ONLY in the following format, with no other text or
explanation:
SCORE: [integer score]
JUSTIFICATION: [brief explanation of the score based on screenshot only]
LINE_ANALYSIS:
[line_number]: [code content] | DESC: [brief description] | IMPACT: [-score]
[line_number]: [code content] | DESC: [brief description] | IMPACT: [-score]

If no significant issues are found, respond with:
SCORE: [integer score]
JUSTIFICATION: [brief explanation]
LINE_ANALYSIS: NONE

[user]
User Prompt: '{user_prompt}'

Generated Code:
```
{code}
```
<rendered screenshot image>
\end{verbatim}
\end{tcolorbox}

\begin{samepage}
\paragraph{Two-Turn Scoring and Line Attribution (2-turn).}
Turn~1 uses the global-only screenshot prompt above, without code, and fixes the score and justification. A separate Turn~2 call receives this assessment, the same screenshot, and the numbered source, and returns only line attributions. The implementation retains the Turn~1 score and does not parse a new score from Turn~2, so source code cannot alter the global visual score. This is the \textit{2-turn} protocol in Appendix~\ref{app:vlm_stability}; it adds one attribution call to global scoring (before retries).

\begin{tcolorbox}[colback=gray!8, colframe=gray!35, arc=4pt, boxrule=0.4pt,
  left=6pt, right=6pt, top=4pt, bottom=4pt, fontupper=\ttfamily\scriptsize]
\begin{verbatim}
[system: turn 2]
You are an expert code reviewer. You have already evaluated a rendered web application
screenshot and produced a score and justification. Now you are given the source code with
line numbers.

Your ONLY task is to identify up to THREE specific code lines that are most responsible for
the issues described in the justification below. Do NOT re-score or change the assessment.

For each problematic line, provide:
- Line number and its code content. IMPORTANT: Each line in the code is prefixed with its
  0-indexed line number (format: "i: code"). When reporting issues, use ONLY these exact line
  numbers.
- Brief description linking the code to the visual issue from the justification
- Impact score (-1 to -3, where -3 indicates severe visual issues)

You MUST respond ONLY in the following format, with no other text or
explanation:
LINE_ANALYSIS:
[line_number]: [code content] | DESC: [brief description] | IMPACT: [-score]
[line_number]: [code content] | DESC: [brief description] | IMPACT: [-score]

If no significant issues are found, respond with:
LINE_ANALYSIS: NONE

[user: turn 2]
User Prompt: '{user_prompt}'

Previous Assessment:
SCORE: {score}
JUSTIFICATION: {justification}

Generated Code:
```
{code}
```
<rendered screenshot image>
\end{verbatim}
\end{tcolorbox}
\end{samepage}

\subsection{Pairwise VLM Judge}
\label{app:pairwise_judge_prompt}

\paragraph{Win-Rate Evaluation.} Given the user prompt and two rendered screenshots (one per model), the judge must pick the better app; ties are disallowed to force a decision. To remove position bias we run a \emph{4-round ABBA} protocol per pair -- the two screenshots are presented in the orders $A,B,B,A$ (each model is shown first in two of the four rounds) -- and take the majority vote; a $2$--$2$ split counts as a tie worth $0.5$ to each side (\S\ref{sec:react_variants}).

\begin{tcolorbox}[colback=gray!8, colframe=gray!35, arc=4pt, boxrule=0.4pt,
  left=6pt, right=6pt, top=4pt, bottom=4pt, fontupper=\ttfamily\scriptsize]
\begin{verbatim}
[system]
You are judging screenshots of single-file React + TypeScript web
applications built with Tailwind CSS.

The user gave a prompt and two models each generated a React app. You are
shown a screenshot of each result.

Pick which screenshot is the better React app. Consider:
- Does it look like a complete, functional React application (not a blank
  page or error)?
- Layout quality: spacing, alignment, visual hierarchy
- Completeness: does it address the user's request with appropriate UI
  components (inputs, buttons, lists, etc.)?
- Visual polish: color usage, typography, overall aesthetics
- Usability: would a user be able to interact with this app?

You MUST pick a winner. Ties are NOT allowed. Even if both are close, pick
the one that is slightly better. Respond with ONLY the single letter: A or B

[user]
User prompt: "{user_query}"

Which screenshot is the better React app? Answer A or B only.
<screenshot A>  Screenshot A (above)
<screenshot B>  Screenshot B (above)
\end{verbatim}
\end{tcolorbox}